\documentclass[sigconf, authorversion, nonacm]{acmart}

\AtBeginDocument{%
  \providecommand\BibTeX{{%
    \normalfont B\kern-0.5em{\scshape i\kern-0.25em b}\kern-0.8em\TeX}}}

\usepackage{cleveref}
\usepackage{subcaption}
\usepackage{wrapfig}
\usepackage{xcolor}

\newcommand{\ourmethod}[0]{{\textit{MaX4Zero}}}
\begin{document}

\title[\ourmethod]{\ourmethod: Masked Extended Attention for \\ Zero-Shot Virtual Try-On In The Wild}

\author{
Nadav Orzech \hspace{0.65cm} 
Yotam Nitzan \hspace{0.65cm} 
Ulysse Mizrahi \hspace{0.65cm} 
Dov Danon \hspace{0.65cm}
Amit H. Bermano \\[0.15cm]
Tel Aviv University 
}

\renewcommand{\shortauthors}{Orzech, et al.}

\begin{abstract}

Virtual Try-On (VTON) is a highly active line of research, with increasing demand.  It aims to replace a piece of garment in an image with one from another, while preserving person and garment characteristics as well as image fidelity. Current literature takes a supervised approach for the task, impairing generalization and imposing heavy computation. In this paper, we present a novel zero-shot training-free method for inpainting a clothing garment by reference. Our approach employs the prior of a diffusion model with no additional training, fully leveraging its native generalization capabilities. The method employs extended attention to transfer image information from reference to target images, overcoming two significant challenges. We first initially warp the reference garment over the target human using deep features, alleviating "texture sticking". We then leverage the extended attention mechanism with careful masking, eliminating leakage of reference background and unwanted influence. Through a user study, qualitative, and quantitative comparison to state-of-the-art approaches, we demonstrate superior image quality and garment preservation compared unseen clothing pieces or human figures.
Code will be available at our project page: \url{https://nadavorzech.github.io/max4zero.github.io/}
 
\end{abstract}

\begin{teaserfigure}
\centering
  \includegraphics[width=1\textwidth]{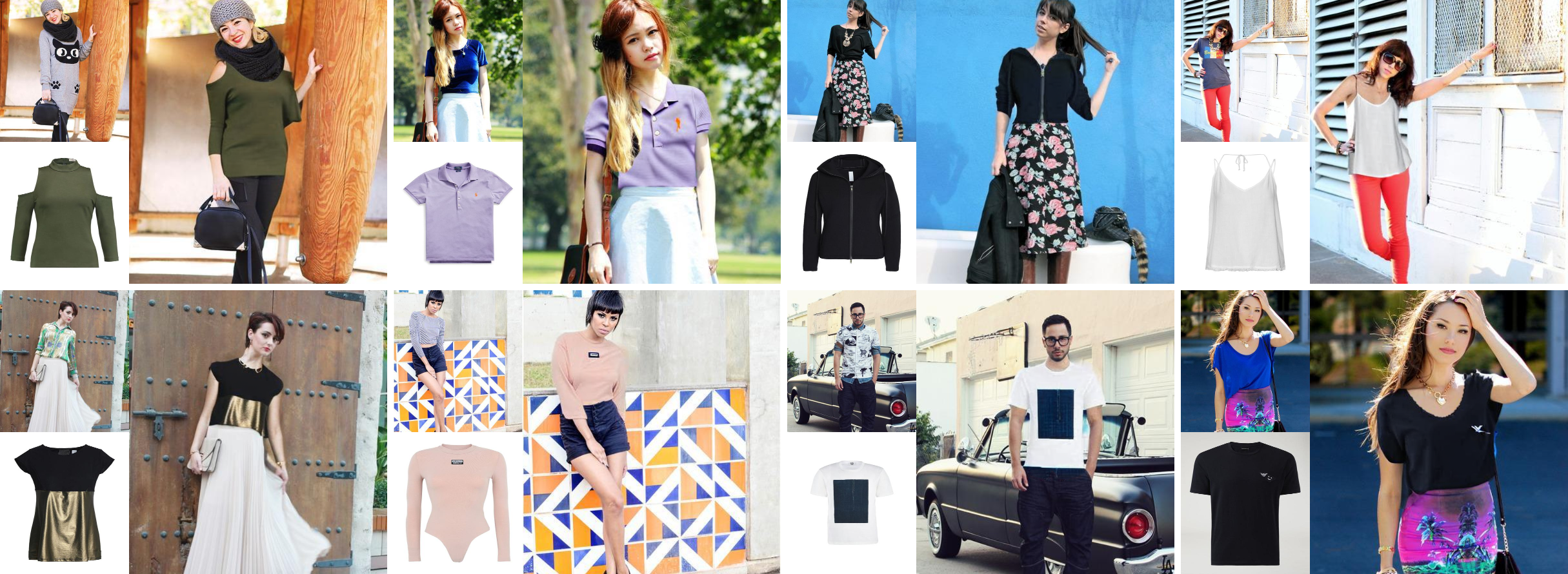}
  \caption{\ourmethod~performs Virtual Try-On in-the-wild (for unseen target images and garments) without any fine-tuning. Given a target image (top), and garment image (bottom), and image is generated using a diffusion-based prior that replaces the input garment with the one already worn in the target (right).}
  \label{fig:teaser}
\end{teaserfigure}
\maketitle

\definecolor{regb}{RGB}{191, 44, 35}
\definecolor{regc}{RGB}{47, 103, 177}
\definecolor{regd}{RGB}{64, 176, 166}

\section{Introduction}
\label{sec:intro}
Current generative models showcase remarkable realism and unprecedented understanding of visual concepts, but at the same time, they lack control. While even a motorcycle riding panda can be generated persuasively, it is difficult to control exactly which jacket the panda should be wearing without drifting away from the natural image manifold. 
A prominent example for where precise identity and high realism are crucial is Virtual Try-On (VTON). VTON, where a garment taken from one image replaces a garment in another, is a highly active field with direct and immediate commercial applications for the fashion, retail and e-commerce industries.
State-of-the-art VTON literature exclusively employs supervision, employing a set of reference-target-result images to fine tune the generative prior to better adhere to the VTON task.
As we demonstrate, this approach does not generalize well to unseen garments, impairing the capabilities of the powerful backbone prior through catastrophic forgetting. 
Instead, we argue contemporary priors already posses the knowledge to handle the task, and only require proper guidance. Hence, we turn to image control literature. Since garment identity is fundamental in our case, common generative based image editing techniques \cite{meng2022sdedit} are unsuitable, as they do not preserve the patterns and nuances of the fabric well, as we demonstrate. Personalization methods \cite{gal2022image, gal2022image2, 10.1145/3588432.3591506, 10.1145/3610548.3618173,Ruiz_2023_CVPR, nitzan2022mystyle, ye2023ip-adapter} fit the task closer, however they typically require per-garment training, which is unfeasible for real-world scenarios. In addition, as we demonstrate, they are inferior in terms of quality for the specific VTON case. 

In this paper we present \ourmethod~ --- the first Zero-Shot In-the-Wild Virtual Try-On approach, realistically and faithfully augmenting unseen reference items without training. 
Following recent personalization and consistency literature, we propose employing Extended Attention \cite{cao2023masactrl, geyer2023tokenflow, alaluf2023crossimage}--- an attention mechanism mixing information between images during the generation process, encouraging data flow between the reference and target images, resulting in semantic similarity between the two.
In doing so, we observe two primary challenges.

First, we see a phenomenon that has been referred to in the literature as \textit{Texture Sticking} \cite{Karras2021}; as strong as the prior is, some details from the reference image are not understood correctly and treated as texture, presenting "sticking" behavior, where position and orientation of features are not determined semantically, but rather from features in the target image. In order to mitigate this, we warp the reference image to match the target in an initial registration step. This encourages the elements to be positioned correctly during transfer. The correspondences for the deformation are extracted from of the same prior, and gaps between the original garment and the deformed one are also inpainted using the prior. 

The second challenge is leakage of background details from the reference garment image, which are clearly irrelevant to the VTON process. To alleviate leakage, we propose masking the attention of unrelated regions in the extended attention, preventing background regions from leaking into the main generation. 
Together, these steps balance the preservation of garment identity and details, while leveraging the prior to ensure the resulting image is on the natural image manifold. 

We evaluate our approach against state-of-the-art VTON \cite{kim2023stableviton, morelli2023ladivton}, image editing \cite{ye2023ip-adapter}, and paint-by-reference method \cite{chen2024anydoor}, and demonstrate, through qualitative and quantitative experiments and a user study, the right balance between identity and garment preservation and resulting realism.  

\section{Related Work}
\label{sec:related}

\subsection{Virtual try-on}
Virtual try-on has a long history of research \cite{6253188}, yet has only known recent leaps in quality through the use of generative models. We here go over some of the most recent developments and state-of-the-art.
TryOnDiffusion \cite{zhu2023tryondiff}, trains a specific "Parallel-UNet", which is able to in parallel diffuse an input garment and an input masked person together with a pose estimate into a low resolution tryon image, which is then upscaled with another standard diffusion model. While transfer quality is very high, the method loses the identity on the target person, as it tends to replace the body entirely and alter details such as tattoos, body shape, etc.

LaDI-VTON \cite{morelli2023ladivton} uses a combination of Textual Inversion \cite{gal2022image} on the garment image, and the feeding of both images to a diffusion model together with "Enhanced Mask-Aware Skip Connections" which directly send inputs the UNet outputs in order to adjust fine details in the final output.

StableVITON \cite{kim2023stableviton}, on the other hand, merges the person and garment images (together with dense pose estimation and masks) into two separate encoders which are then merged into the UNet decoder, using the garment encoding as keys and values for attention, while person encodings are used as queries. The technique additionally introduces a new penalty term to the diffusion loss based on attention map total variation, aiming at further preserving garment identity through sharper attention maps.

\begin{figure*}[ht]
    \centering
    \includegraphics[width=0.99\textwidth]{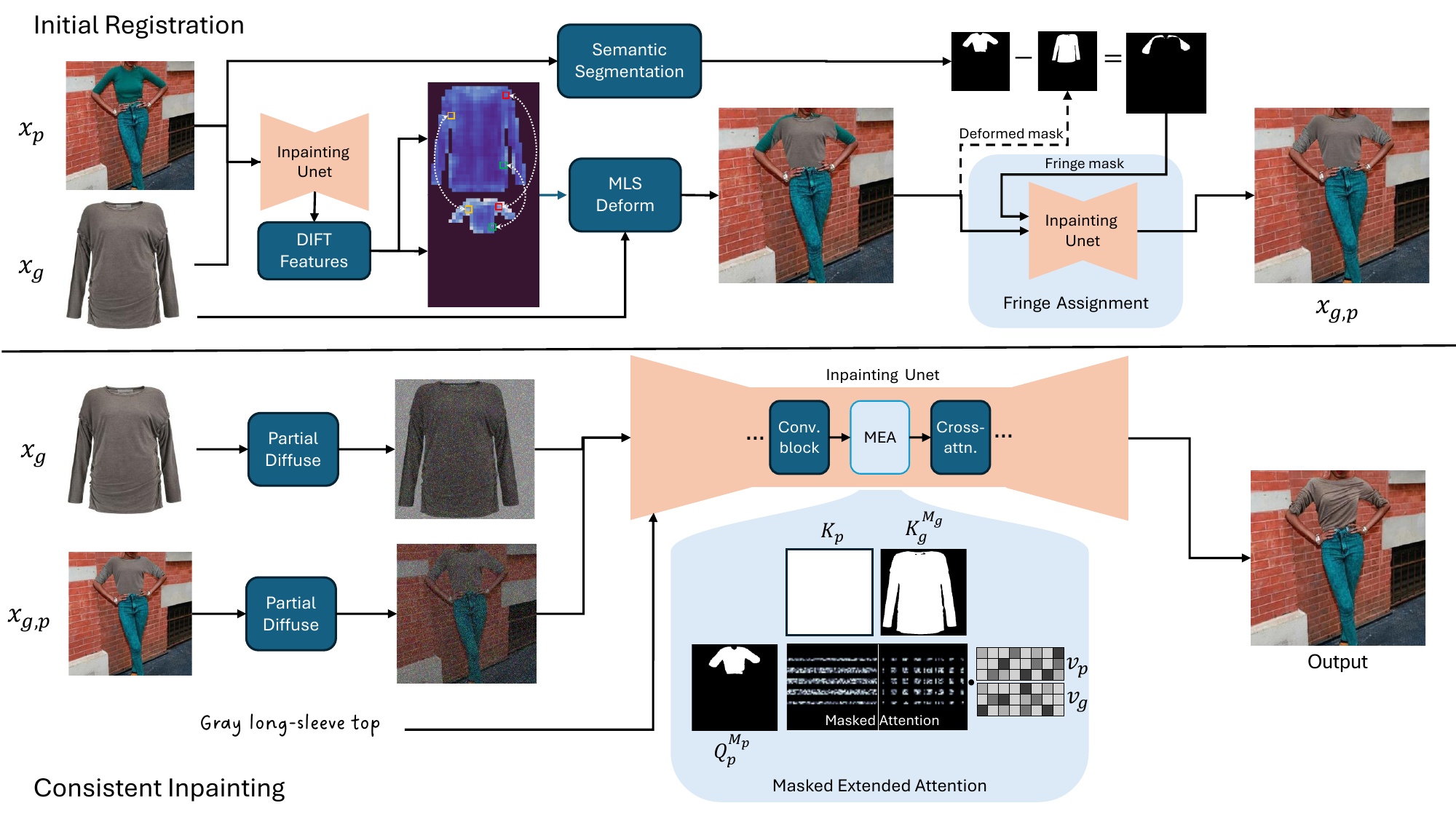}
    \caption{Overview of the proposed \ourmethod~ method. \textbf{Top:} the \textit{Initial Registration} stage, where the reference garment is warped to match the target person using extracted deep features from both images \cite{tang2023emergent}. The remaining gaps between the target garment and warped one are filled by the Fringe Assignment module (see \Cref{fig:fringe}). \textbf{Bottom:} the \textit{Consistent Inpainting} stage, where utilizing the Masked Extended Attention mechanism for transferring the reference fine-details through stroke-based inpainting.}
    \label{fig:overview}
    
\end{figure*}

\subsection{Image editing with diffusion models}

Diffusion models \cite{ho2020denoising, song2022denoising, nichol2022glide, ramesh2022hierarchical, rombach2022highresolution} have rapidly advanced the field of image editing, demonstrating remarkable flexibility and quality across various applications. Perhaps the simplest technique for image editing is  \emph{inpainting} 
\cite{lugmayr2022repaint,wang2023imagen,xie2023smartbrush,nitzan2024lazy, rombach2022high}
-- the task of completing a masked, or missing, image region. 

In the context of modern diffusion models inpainting is performed by running the denoising process conditioned on the non-masked parts of the image and a text prompt

This approach is restricted to local edits, as only the masked region may be modified, and cannot reuse information contained within the mask.

Another angle to perform whole-image editing revolves around inversion techniques. 
Given a diffusion model and an image, inversion refers to the process of recovering the input noise which would lead to the generation of original image after denoising.
These methods, usually referred to under the umbrella term of DDIM Inversion \cite{song2022denoising, dhariwal2021diffusion}, require careful selection of the renoising process in order to recover truly inverted noise, with examples such as Null-text Inversion \cite{mokady2022nti}, LEdits++ \cite{brack2023leditspp} or ReNoise \cite{garibi2024renoise}. Some also perform inversion directly in image-space rather than latent-space, as seen in \cite{huberman2023ddpminv}. 
All these methods primarily utilize global editing techniques. However, for the Virtual Try-On task, maintaining the integrity of the background is paramount.

Another avenue for editing is that of object transfer, i.e. editing an image by adding or replacing objects in the image using a source image containing the objects. This by nature preserves the identity of the transfered object entirely. A recent technique showing great promise, Anydoor \cite{chen2024anydoor}, achieves this by both encoding the source object through an identity encoder, and merging several levels of detail frequencies to obtain perturbations which are fed into the UNet activations.

\subsection{Identity-preserving generation}

The topic of preserving the identity of a subject during generation is a natural one when generating images or videos where a specific subject is desired to be seen. Typically, applications have revolved around preserving the identity of people, in particular facial features.

The most naive way to perform identity preservation is, given a collection of images representing a subject, to fine-tune a pre-trained diffusion model using said collection \cite{nitzan2022mystyle,Ruiz_2023_CVPR}. As this is a costly process, many techniques have been developed to avoid training on the entire model, such as LoRA \cite{hu2021lora}, FedPara \cite{woo2021fedpara}, OFT \cite{qiu2023oft}, CustomDiffusion \cite{kumari2022customdiff}, CONES \cite{liu2023cones} which to train low parameter count perturbations or transformations of the original weights. The main downside is that they require many images of the subject, and may be very limited in the scope of successful generations.

A more advanced way of preserving identity is the use of a pretrained identity-preserving module which tweaks the diffusion model's weights or activations conditional to an identity-providing image. One of the most popular, IP-adapter \cite{ye2023ip-adapter}, uses a trained decoupled cross-attention module which injects perturbations derived from an identity image into the diffusion UNet in order to condition the diffusion, much like in AnyDoor (see above). IP-adapter is however pretrained with a strong emphasis on the preservation of the identity of human faces, and struggles with other types of conditionings, especially garments. While overall appearance is preserved, most details do not transfer correctly.

\section{Method}

Given a person image $x_p\in \mathbb{R}^{H_p\times W_p\times 3}$ and a garment image $x_g\in \mathbb{R}^{H_g\times W_g\times 3}$, our goal is to synthesize the garment object on top of the person, while removing previous existing garment and preserving all other details. 
In this work we address the task of virtual-try on through a reference-based image inpainting approach. In other words, we inpaint the area within the previous garment region of the person's image $x_p$, using the desired garment as reference $x_g$.
In order to perform this inpainting, we utilize a pre-trained text-to-image model (Stable Diffusion \cite{rombach2022high} in our case). Our zero-shot method \ourmethod, is composed of 2 stages as presented in Fig \ref{fig:overview} - (1) \textit{Initial Registration} where the garment is placed on top of the person while filling the created fringes, providing powerful ques for fine generation (2) \textit{Reference based Consistent Inpainting} for inserting fine-grained details and increasing fidelity. 

\subsection{Preliminaries}

\textit{Stable Diffusion} is an open source diffusion model variant which initiates its denoising process in a latent space. Each image is first fed forward through a pretrained encoder, transforming it into the latent space, and in the final step of the process a pretrained decoder maps the generated output back to RGB space. 
At each timestep, the encoded input undergoes gradual denoising within the stable diffusion model's U-Net architecture, which incorporates cross-attention and self-attention blocks.
The self-attention blocks focus on image details and cross-attention blocks are utilized to incorporate text prompts.

In each \textit{self-attention} block, the intermediate features are projected into queries $Q$, keys $K$, and values $V$ with dimension $d$. 
Each self attention layer first calculates similarity scores between all $Q$ and $K$ vectors. 
Then the normalized values vectors are projected on the attention matrix creating the block output $\phi$ as a weighted sum of the values vectors. Formally:
$$A=softmax\left(\frac{Q\cdot K^\top}{\sqrt{d}}\right)$$
$$\phi=A\cdot V$$

where $A$ is the computed attention map and $\phi$ is the block output. This process is applied to each query independently allowing non-restricted receptive field by capturing correspondences across the entire image.

\subsection{Initial Registration}
\begin{figure}[h]
    \centering
    \includegraphics[width=0.5\textwidth]{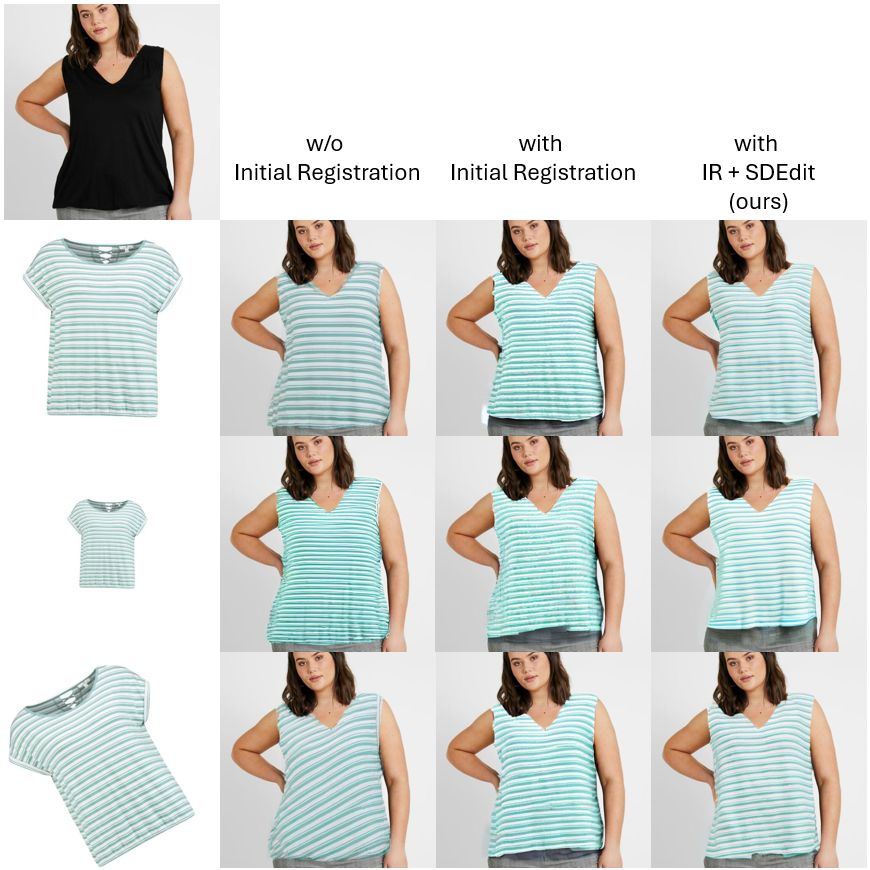}
    \caption{Garment generation compared including or excluding the initial registration stage. This demonstrates the "texture sticking" phenomenon,  where the spatial features of the transformed shirt are apparent in the resulting generation.}
    \label{fig:wrapping}
\end{figure}

In order to faithfully inpaint the given garment on top of the target person, transferring the garment texture or appearance is not enough, as presented in Fig \ref{fig:wrapping}. As can be seen, we observe that explicit spatial features in $x_g$, sensitive to scales, rotations, and even translations, can be found in the generated result, even though they carry no semantic meaning. We postulate this is because some features cannot be fully avoided or translated due to limited capacity of the prior, in a phenomena similar to that called \textit{Texture Sticking} in the literature \cite{Karras2021}.
We hence first align the garment on the target image and relevant masked area. this task becomes more complicated when working with real world images in contrary to generated ones, since we don't have control on the generated image layout. 

\textbf{Garment Warping.} 
To overcome this issue, we develop a deep features based deformation block inspired by \cite{tang2023emergent}. In the warping phase, we initially employ the forward process of the diffusion model, by adding noise corresponding to timestep $t$ to both images. Then, we pass the images through the denoising process, while extracting the deep features from the model's transformer blocks of both the target previous garment and the required referenced garment. According to the extracted features, we can create a bi-linear correspondence matrix between the two garments by identifying the best semantic matches, which referred to as two vectors exhibiting the minimum distance between them. This correspondence matrix is then used in order to deform the reference garment on top of the target person using a non-rigid deformation scheme named Moving-Least-Squares (MLS) \cite{10.1145/1179352.1141920}. 

\begin{figure}[h] 
    \centering
    \includegraphics[width=0.9\columnwidth]{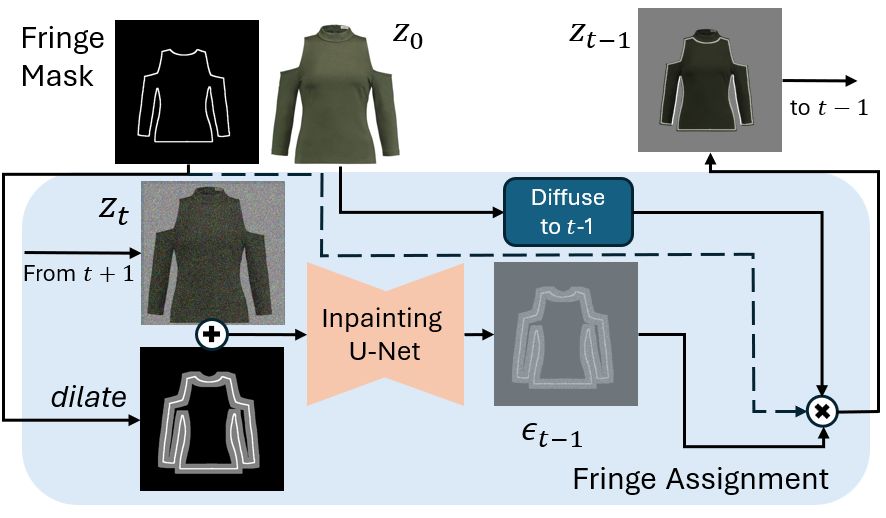}
    \caption{Overview of the double-mask inpainting strategy. At each timestep, the dilated mask is used for noise prediction, but only the regions in the thin mask are actually taken to the next diffusion iteration.}
    \label{fig:fringe}
\end{figure}

\textbf{Fringe Assignment.}
After warping the reference garment there are, however, some fringes left unwrapped because the two garments are not identical, and the posture of the target person is arbitrary. 
We fill these fringes in by applying inpainting, asking to complete the regions left by the old garment and are not covered by the current one. 
Previous work \cite{Avrahami_2023} shows that inpainting models fail to properly alter small masks. Thus we chose to use a double-mask strategy, consisting of the original fringe mask and a dilated one.
As depicted in Fig \ref{fig:fringe}, at timestep $t$ of the inpainting process we use the dilated mask to predict the defused noise $\epsilon_{t-1}$. However, the iteration output $z_{t-1}$ is constructed from the diffused image $z_{t-1}$, where the model prediction is applied only to the original thin fringe mask. 

Our experiments show that by using the double-masks strategy we benefit from both worlds - the diffusion model is able to generate valid information on one hand, and on the other, the information still complies with the background seamlessly.

\subsection{Masked Extended Attention (MEA)}
Once we have the coarse features of the garment on top of the person image, placed in the correct location, we can turn to fine detail enhancement. Previous works \cite{cao2023masactrl, geyer2023tokenflow, alaluf2023crossimage} explored the self-attention layers within the denoising network of a text-to-image diffusion model. More specifically, it has been demonstrated that it is possible to utilize the queries, keys, and values from these self-attention layers in order to transfer semantic information between different and unrelated images. 

\textbf{Masked Extended Attention (MEA).}
Prior research \cite{alaluf2023crossimage} has shown that queries control the semantic meaning of each spatial location, while the keys provide contextual information for each query. This enables the model to assess the significance of various image segments for a given query position by multiplying the relevant query and key. Then, the values represent the content to be generated and specify the information used to determine the relevant features of each query position.

Given those observations, and inspired by \cite{cao2023masactrl, geyer2023tokenflow}, our approach as demonstrated in Fig. \ref{fig:overview} is to expand the self-attention keys and values during the generation of the target image, allowing it to include information from the reference image, yet the queries remains fixed. 
In other words, during generation of the target garment, we allow the new generated garment values to be determined not only by the same image context, but also from the garment example. 

We observe that using this extended attention results in background information leakage from the reference garment into the target result. To overcome this challenge, inspired by \cite{cao2023masactrl}, we carefully masked the relevant parts within the attention masks. The use of masks as part of the extended attention mechanism allows us to establish control on which information propagate between the images and through different patches within the same image. We named this mechanism Masked Extend Attention or MEA. 

In more details, the input to the MEA block are the self-attention features from both images. Namely, the queries, keys and values from both the target person image $Q_p, K_p, V_p$, and the reference garment $Q_g, K_g, V_g$. As mention, we expand the input of the extended attention block by utilize the inpainting masks $M_p$ and $M_g$ through the process. The keys of both images are concatenated and the masked extended attention defined as: 
\begin{multline}
  MEA(Q_p, [K_p,K_g], M_p, M_g) = \\
  softmax\left(\frac{(1-M_p)\cdot f^{bg} + M_p \cdot f^{fg}}{\sqrt{d}} \right) 
\end{multline}
where 
$$f^{bg} = Q_p \cdot K_p^\top$$
$$f^{fg} = Q_p \cdot [K_p, M_g\cdot K_g]^\top$$

The block output for target person generation is defined as:
$$\phi^{MEA} = MEA(Q_p, [K_p,K_g], M_p, M_g) \cdot [V_p,V_g] $$

Intuitively, generation of a target garment pixel takes into account the relevant information from the reference image, as well as information from the same generated garment with the scene context. Note that the Masked Extended attention processes the original and un-deformed garment, in order to transfer the authentic and an-distorted attributes to their grounded place.

\textbf{Details Propagation Enhancement.} 
Prior research \cite{alaluf2023crossimage} has shown that conventional self-attention blocks tend to concentrate their attention on a concentrated region around the image patch. 
In contrast, the masked extended attention block distributes its focus across two images as we expand its receptive field beyond just the target image. This results in a more unified attention maps as a softmax is applied to a larger set of features. Thus, this can also introduce noisy information into the block and potentially overlook the desired finer details.
In order to encourage the attention maps to concentrate its focus on more specific regions within the reference image, following \cite{alaluf2023crossimage} we used a contrast operation to enhance the variability of the attention maps where we amplify the emphasis on the peaks of the attention map, effectively nullifying any noisy information. The contrast operator defined by \cite{alaluf2023crossimage} as:
$$Enhance(A)=\left(A-\mu(A)\right)\beta +\mu (A)$$ 
where $\mu$ is a mean operation and $\beta$ is the contrast operator

\textbf{Stroke-Based Inpainting}
Finally, we employed the stroke-based inpainting concept (SDEdit), as introduced in \cite{meng2022sdedit}. By using the registered garment as a initial image, we denoise the image using classifier-free guidance (CFG) \cite{ho2022classifierfree}. At each denoising timestep, we use two parallel forward passes through the denoising network. 
The first pass using the conventional self-attention layers of the network to enhance the realism of the generated garment while preserving the context received from the garment description prompt, yielding in $z^{base}$ and $z^{text}$ respectively. Meanwhile, the second pass leverages our Masked Extended Attention layers to reproduce the reference garment by capturing its fine details, yielding in $z^{MEA}$.
By CFG scheme the predicted noise defined as:  $$z_{t-1}=z^{base}_t+\alpha_{MEA}\cdot(z^{MEA}_t - z^{base}_t) +\alpha_{text}\cdot(z^{text}_t - z^{base}_t)$$

\section{Experimental Results}
\begin{figure*}
    \centering
    \includegraphics[width=0.9\textwidth]{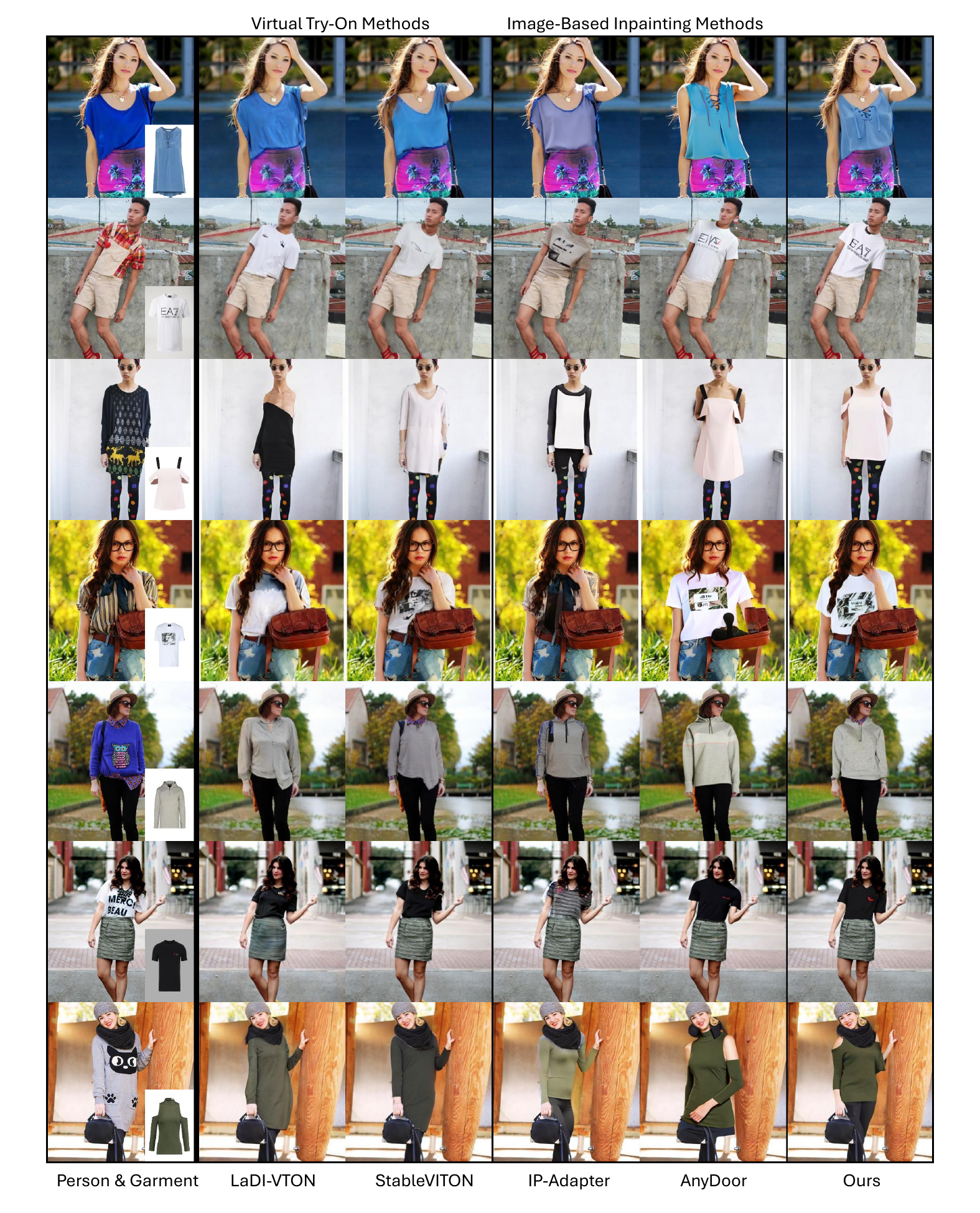}
    \caption{
    Qualitative comparison of \ourmethod~ and competitors on in the wild target images. We compare against LaDI-VTON \cite{morelli2023ladivton} and stableVTON \cite{kim2023stableviton}, which are dedicated VTON approaches, and IP-adapter \cite{ye2023ip-adapter} and Anydoor \cite{chen2024anydoor}, which are personalized image editing and paint-by-reference methods respectively. As can be seen, garment identity is preserved better using our method for unseen garments.}
    \label{fig:qualy}
\end{figure*}

\textbf{Datasets.}
We conduct our experiments using three publicly available datasets. The clothing garment images taken from the virtual-try on datasets DressCode \cite{morelli2022dresscode} and VITON-HD \cite{choi2021viton}, while the description for each garment extract from the work of \cite{baldrati2023multimodal}. The target person images were taken from the work of \cite{Liang_2015}, a large dataset for human in-the-wild parsing. 
To ensure accurate utilization of our benchmarks, we rigorously followed the pre-processing guidelines of VITON-HD dataset to obtain the necessary input conditions.

\textbf{Baselines.}
We compare our method with two diffusion-based virtual try-on methods StableVITON \cite{kim2023stableviton} and LaDI-VTON \cite{morelli2023ladi}, both trained on target person - clothing garment pairs extracted from DressCode and VITON-HD. 
We additionally assessed our method against two example-based models: IP-adapter \cite{ye2023ip-adapter} as a mask oriented image editing technique, and AnyDoor \cite{chen2024anydoor} an inpaint-by-reference method. Both of these methods are extensively trained on large datasets containing diverse objects, with the objective of effectively reconstructing the object within the scene.

\textbf{Evaluation Metrics.}
In order to quantify our method performance, we initially conducted a user study survey with 57 participants. Each participant was asked to assess image coherence, as well as the preservation of both the clothing garment and the target person's identity. Moreover we used Frenchet Inception Distance (FID) score \cite{heusel2018gans} to measure the fidelity of the generated images, and in addition LPIPS \cite{zhang2018unreasonable} and DreamSim \cite{fu2023dreamsim} for evaluating garment attributes transfer and person identity preservation.

\subsection{Qualitative Results}
\begin{figure}[ht]
    \centering
    \includegraphics[width=0.49\textwidth]{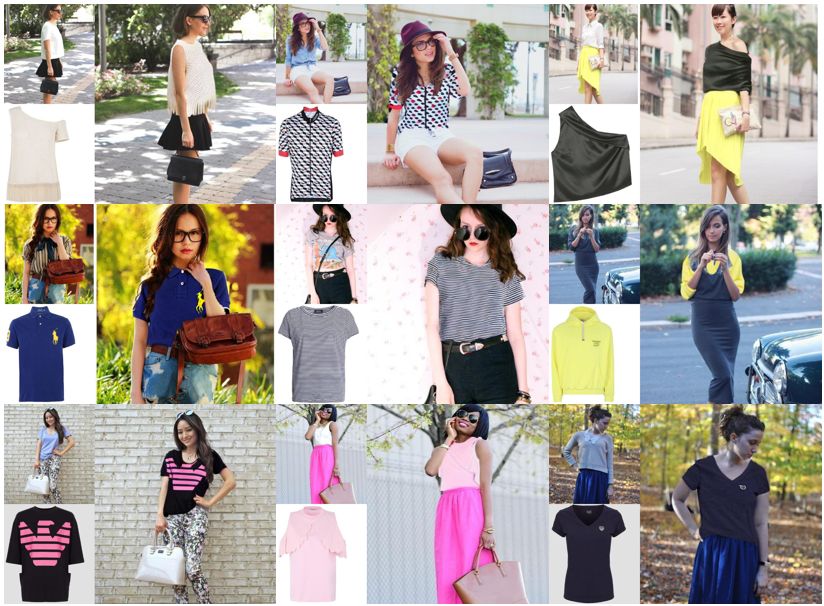}
    \caption{Results Gallery. Generated results by \ourmethod. Best viewed when zoomed in.}
    \label{fig:gallery}
\end{figure}

As we demonstrate in Fig. \ref{fig:qualy}, \ourmethod~ generates realistic images, as the image background remains unchanged due to the advantages of the inpainting scheme. Moreover, \ourmethod~ successfully maintains the in-the-wild target person's identity while accurately transferring the texture and characteristics of the referenced garment, compared to the four baseline methods. Specifically, Anydoor preserve the semantic features of the reference garment, but misses the rotation or spatial location of those, as could be seen on second row. The second image-based inpainting method, IP-adapter able to inpaint a garment which fits the semantic meaning of the requested reference but misses the fine details of it.
On the other hand, virtual try-on methods such as StableVITON and LaDI-VTON which been trained on virtual try-on specific domain are able to create fit and harmonized clothing and appropriately positioned in most cases, but on the expense of failing to capture the characteristics of the garment and transfer it to the arbitrary in-the-wild person, as can be seen in the left two columns. Moreover, in contrary to \ourmethod, those benchmarks alter the background of the target garment or the person posture as can be noticed in all three rows.\\
Additional outcomes produced by \ourmethod~ are depicted in Figures \ref{fig:teaser}, \ref{fig:gallery} and \ref{fig:results_grid}. 
Furthermore, \ourmethod~ method allow us to try-on out of domain reference garments or target figure, as depict in Figures \ref{fig:results_grid_ood_refs} and \ref{fig:results_grid_ood_tars}.
See the supplementary material for more qualitative results.

\subsection{Quantitative Results}

\begin{table}[ht]
    \centering
    \caption{Quantitative comparison for in the wild settings. Results in \textbf{Bold} and \underline{Underlined} refers to best and second best results, respectively. We measure garment identity using two semantic similarity methods (LPIPS \cite{zhang2018unreasonable} and DreamSim (DrSm) \cite{fu2023dreamsim}) on the garment region of the image alone, person preservation using the same methods on the background region, and overall realism using FID \cite{heusel2018gans} scores. For all metrics, lower is better. As can be seen, our approach offers a balance between realism and identity preservation, supporting users' preferences in the user study.}
    \begin{tabular}{c | c c | c c | c }
        \toprule
        & \multicolumn{2}{c|}{Garment ID} & \multicolumn{2}{c|}{Person Pres.}  \\
        {$Method$} & {$LPIPS$} & {$DrSm$} & {$LPIPS$} & {$DrSm$} & {$FID$}\\
        \midrule    
        LaDI-VTON       & 0.679 & 0.619 & 0.146 & 0.047 &  74.78  \\ 
        Stable-VITON    & 0.686 & 0.639 & 0.071 & 0.013 &  67.56  \\
        \midrule    
        IP-Adapter      & 0.687 & 0.626 & \textbf{0.016} & \textbf{0.001 } & \textbf{64.33} \\
        AnyDoor         & \textbf{0.619} & \textbf{0.434} & 0.059 & 0.059& 76.80   \\
        \midrule 
        Ours            & \underline{0.643} & \underline{0.518} & \underline{0.021} & \underline{0.009} & \underline{66.27}
    \end{tabular}
    \label{tab:quant_results}
\end{table}

\textbf{Metrics.}
In effort to properly demonstrate the success of each method we used several evaluation metrics. First, to evaluate properly reference garment transfer we used LPIPS and DreamSim as similarity scores, between the generated garment and the referenced one. Using the same reasoning, we addressed the person identity preservation by LPIPS \cite{zhang2018unreasonable} and DreamSim \cite{fu2023dreamsim} again as similarity score, but in this case we evaluated the generated image with the clothing garment cropped out against the original person figure image. Lastly, we used FID \cite{heusel2018gans} score between the generated images and the original in-the-wild human images in order to evaluate the fidelity and realism of the generated images. 

\textbf{In The Wild Evaluation.} 
Our objective here is to assess our method against benchmark models using arbitrary target person images, which may vary in terms of posture and background. This is in contrast to the VITON-HD and DressCode datasets, where the backgrounds are constant and the human figures have similar poses.
As shown in Table \ref{tab:quant_results}, regarding properly transfer garment identity, our training-free method ranked as the second best between the fine tuned methods closely trailing behind AnyDoor, according to the explained similarity metrics. The relative succession of AnyDoor in this term could be attributed to the extensive resources that AnyDoor allocates to acquire reference details as presented in \cite{chen2024anydoor}. However, AnyDoor exhibited inferior results in person and background preservation compared to our method and other benchmarks. 
In terms of preserving the person and background, our method secured the second-best position, with a narrow margin behind IP-Adapter, as both first methods using the inpainting approach. 
Moreover, the results presented in Table \ref{tab:quant_results}, demonstrate similar trend regarding overall image fidelity, as indicated by the FID scores.
On the other hand, IP-adapter struggled to accurately transfer the garment attributes, as it ranked near the bottom according to the garment similarity metrics. 
Note that both pretrained virtual try-on method did not evaluated as top 2 methods in both metrics.
In overall perspective of both tasks, our training free and zero-shot method was found comparable to the task-specific fine-tuned models.

\begin{figure}[ht]
    \centering
    \includegraphics[width=0.5\textwidth]{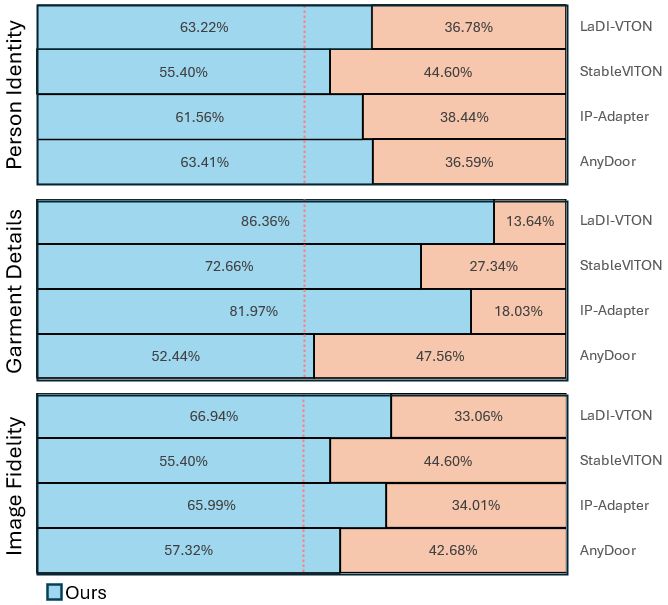}
    \caption{User Study results, comparing three aspects of \ourmethod~results against four benchmark methods. Based on 57 participants.}
    \label{fig:user_study}
\end{figure}

\textbf{User Study.}
In order to quantify further our results against the other methods on in-the-wild target person images we conducted a user study involving 57 participants. 
Each participant was presented with the target person image, the reference garment, and two results: one generated by our model and the other by a benchmark method. 
Then each one was asked to determine which result excelled in three aspects: (1) Preserving the target person identity and posture, (2) Accurately capturing garment attributes, and (3) Overall image fidelity. As depicted in Fig. \ref{fig:user_study}, our method surpassed all other fine-tuned benchmarks across all three aspects.
As shown in \ref{fig:user_study}, the comparison in assessing garment attributes was closely contested when compared with AnyDoor. 
Furthermore, the StableVITON method proves formidable in terms of preserving the identity of the person and overall fidelity, owing to its capability to maintain the original features of the image.
Note that the trend observed in our user study resembles to the quantitative results presented earlier. However, when comparing the two results side by side, our method was found to be outperformed in human perception.

\subsection{Ablation Study}
In order to demonstrate the effectiveness of our method components we evaluated the effect of each component on the final result. 

\textbf{Initial Registration.}
First, we investigated the effectiveness of the wrapping module in the initial registration stage and how it influences the final results. 
In Fig. \ref{fig:wrapping}, we visualize the impact on the final garment, both with and without initial registration, using the same reference clothing piece with different augmentations. In both cases we inpainted the required masked area of the target image, while addressing each augmented reference garment or deformed version of it. As can be seen in the figure, when neglecting the initial registration stage, the augmentation on the reference image effects the generated image, whether it is a more dense texture when using small-scale garment in the second row, or propagation of tilted texture as in the third row. However, this issue is solved when using the initial registration stage, as the deformation of the reference aligns the attributes to the same proper location in all cases.
We believe that this phenomenon occurs because the MEA, even though is an attention mechanism, still operates on the patch level, giving significant bias to immediate surroundings..

\begin{figure}[ht]
    \centering
    \includegraphics[width=0.5\textwidth]{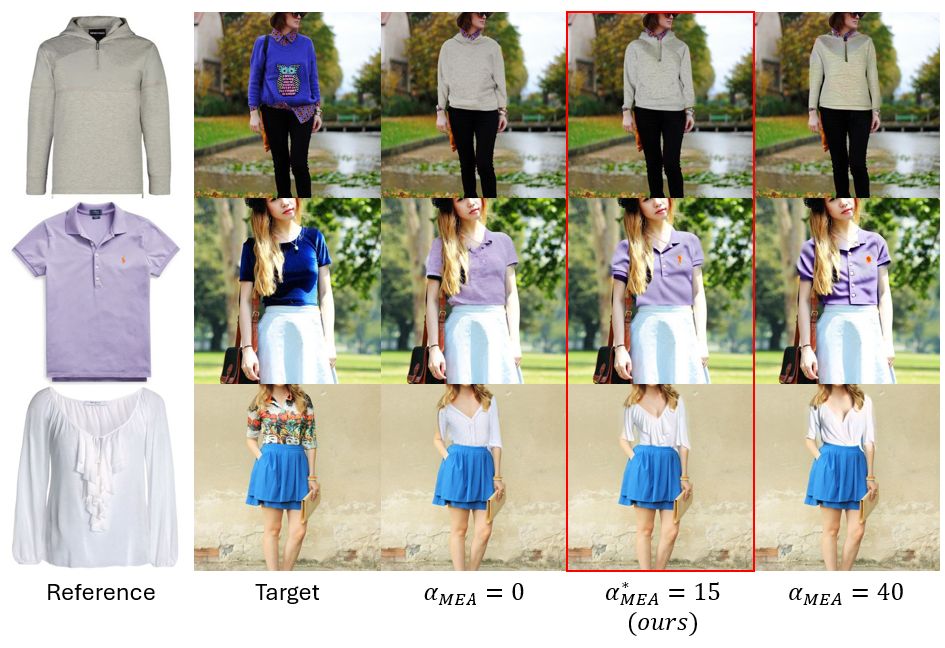}
    \caption{Ablation study of the \textit{Consistent Inpainting} stage using different MEA guidance factors. We set $\alpha_{MEA}=15$ for all our experiments}
    \label{fig:mea_ablation}
\end{figure}

\textbf{Extended Attention}
The effectiveness of the MEA mechanism is showcased in Fig \ref{fig:mea_ablation}. Here, we explore the impact of varying the MEA guidance scale, ranging from $\alpha_{MEA}=0$ when the mechanism is inactive, through our selected hyperparameter $\alpha_{MEA}=15$, to an exceptionally high value of $\alpha_{MEA}=40$.

In the left column, where the MEA mechanism is inactive, it can be noticed that the fine-details of the reference garment fail to extend into the target image. This is due to the fact that during inpainting the inpainted area lacks contextual information from beyond the target image boundaries. Meaning, the outcome is stroke-based inpainting of the warped garment area, utilizing the inpainting model prior knowledge alone.
On the other hand, as shown on the right column, when increasing excessively the MEA guidance factor, fine-grained details over propagates into the target image, whether it reflected in extra buttons as seen in the second row, or generation of shelf-like garment, disrupting the harmonized target image.

\begin{figure}[ht]
    \centering
    \includegraphics[width=0.5\textwidth]{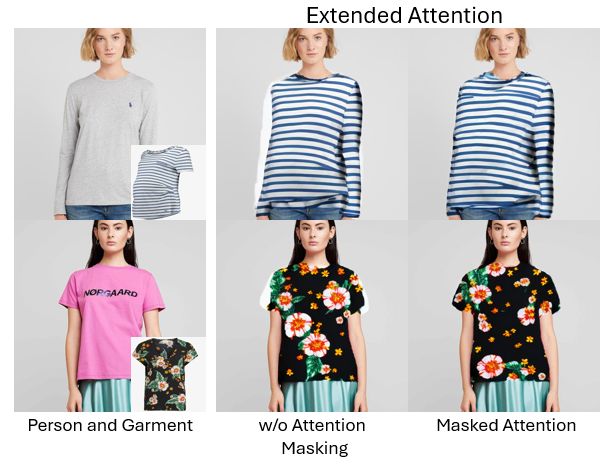}
    \caption{Ablation study on the masking of the Extended Attention mechanism. As can be seen, masking mitigates the leakage of the white background into the generation. }
    \label{fig:mask_ablation}
\end{figure}

\textbf{Extended Attention with Masking (MEA)}
We highlighted the significant of employing masks in the MEA mechanism in Fig \ref{fig:mask_ablation}. As discussed earlier, applying masks to the extended attention process enables us to regulate the flow of information between the reference and target images. In the left column, we observe background elements from the reference image seeping into the inpainted segments. This results in two distinct artifacts: firstly, the emergence of white patches along the generated garment boundary, as evident in the first row; secondly, an increase in the background color's influence within the generated texture of the clothing piece relative to the reference image, as reflected in more notable white flowers in the second row.

\section{Discussion}
In this work we introduced a novel zero-shot in-the-wild approach to the image-based Virtual Try-On task. 
Our purpose method comprises two stages. Initially we ground the coarse feature of the reference garment onto the target person's body by wrapping the garment around it and filling in any gaps created by incomplete matching. As shown in our work, this stage provides strong cues for the next step, ensuring the right features are generated in the right positions, diminishing the bias originating from the reference garment image.
Once we have the grounded garment, we introduce a Masked Extended Attention mechanism. This mechanism pours the fine details from the reference onto the grounded location on the target through an inpainting diffusion process.
In our experiments we demonstrated that our method outperforms other supervised-trained virtual-try-on benchmarks on in-the-wild human images. Furthermore, we emphasized the critical role of each component in contributing to the success of our entire pipeline.

\textbf{Societal Impact.} Misuse of this technology, and other virtual try-on techniques, include disinformation, fake news, and body-related embarrassment on public channels such as social media. Our work contributes to the democratization of VTON tools, potentially making them more convincing and approachable. Viable solutions include model watermarking and real-time detection on public platforms. We call potential users of our code not to cross the line of causing discomfort or violating national regulations, and the research community to continue developing in-place reliable detection methods.

We hope our work will contribute to resolving the virtual-try-on task with great generality, speed, and resource demands. We encourage further exploration of these tuning free zero-shot two-stage methods to tackle image-based inpainting across various domains.

\begin{figure}[h]
    \centering
    \includegraphics[width=0.49\textwidth]{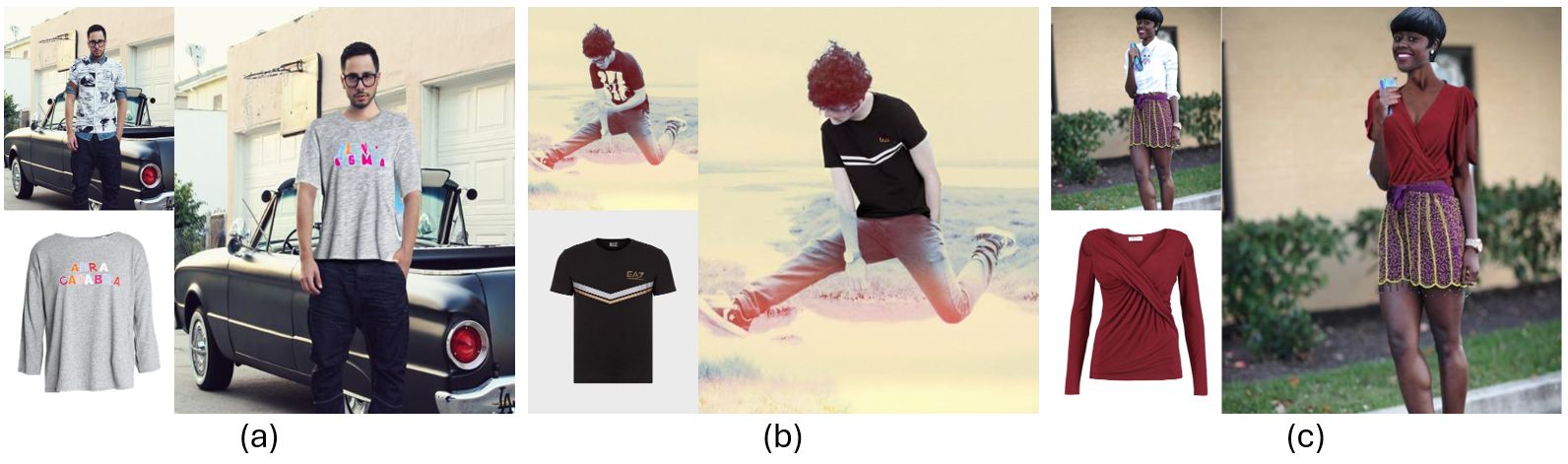}
    \caption{Limitations: Our prior still suffers when reconstructing text and elaborate prints (a) and is sensitive to mask placement as it might alter object within the mask (b),(c).}
    \label{fig:limit}
\end{figure}

\subsection{Limitations}
Current Stable Diffusion-based architectures still exhibit shortcomings in accurately reproducing readable and coherent textual elements such as text or logos. In some fail cases our model is able to reproduce the overall looks or fine details of the textual object, but struggles to define precise and highly comprehensible letters or numbers, as shown in Fig \ref{fig:limit}(a).
Another issue our model struggle with is the sensitivity for the mask placement. As can be seen in Fig \ref{fig:limit}(b), once the mask includes other human body segments (in this case the left hand), our method might alter it position or looks, fitting it to an overall authentic image, however not identical to the original one. For the same reason, it is evident from Fig \ref{fig:limit}(c) that when the inpainting mask extends beyond its intended boundaries into other objects, such as hair, palms, or in this instance, a book held by the human figure, our method might alter it, resulting in disruptions to the original image. An interesting avenue for future research is to accommodate this mask in a more semantic or latent manner, rather then through explicit spatial arrangement.  

\section{Acknowledgements}
This work was supported in part by the Israel Science Foundation under Grant No. 1337/22.

\bibliographystyle{ACM-Reference-Format}
\bibliography{sigbib}

\clearpage

\appendix
{\bf {\LARGE APPENDIX}}

\section{Implementation Details}
In our work, we adopted the pipeline of Stable Diffusion v2 using the default hyperparameters and input dimensions of $512\times512$. For non-square images we cropped out the relevant masked area and patch it back at the end of the process. For the inpainting process, we employ the standard DDIM scheduler for 50 denoising steps.
\textbf{Garment Warping -} we extracted the deep features from the second layer of the decoder part of the denoising U-Net. For the the deformation we used the bi-linear correspondence matrix, while omitting outlier matches which cause distortions in the warped garment.
\textbf{Masked Extended Attention -} For the second stage, we replace the conventional
self-attention layers with our MEA layers within the denoising U-Net only in the decoder part and at attention size of $32\times32$ and $64\times64$ during the entire denoising process. 
\textbf{Details Propagation Enhancement -} we chose to enhance our masked extended attention maps using contrast factor of $\beta=1.5$. 
\textbf{Stroke-Based Inpainting -} the stroke based phase using the MEA mechanism set to the last $35\%$ of the denoising process.

\section{Additional Qualitative Results}
\begin{figure*}[ht]
    \centering
    \includegraphics[width=0.85\textwidth]{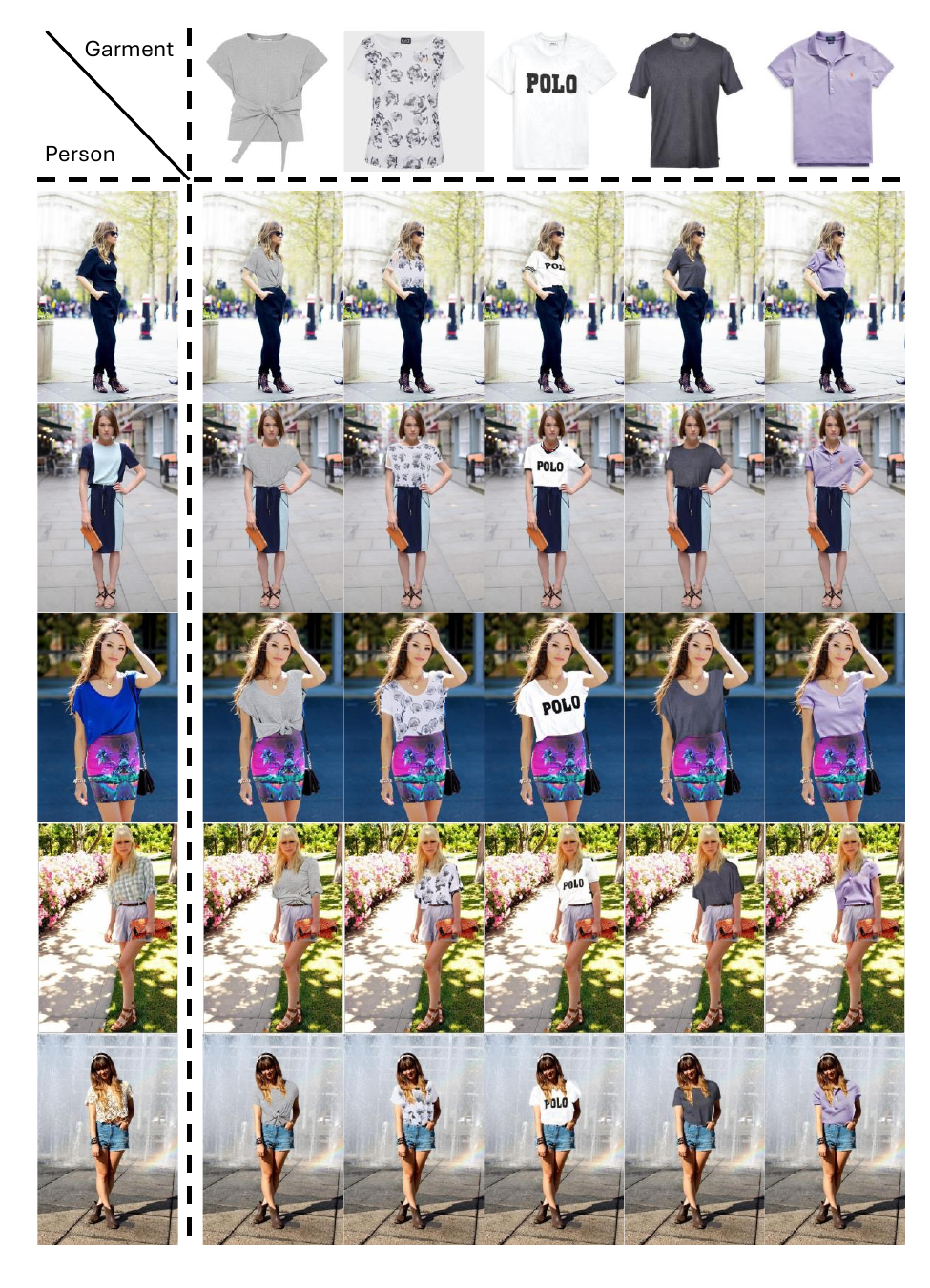}
    \caption{
    In-the-wild virtual-try-on results generated by \ourmethod. Best viewed when zoomed in.
    }
    \label{fig:results_grid}
\end{figure*}

\begin{figure*}[ht]
    \centering
    \includegraphics[width=0.85\textwidth]{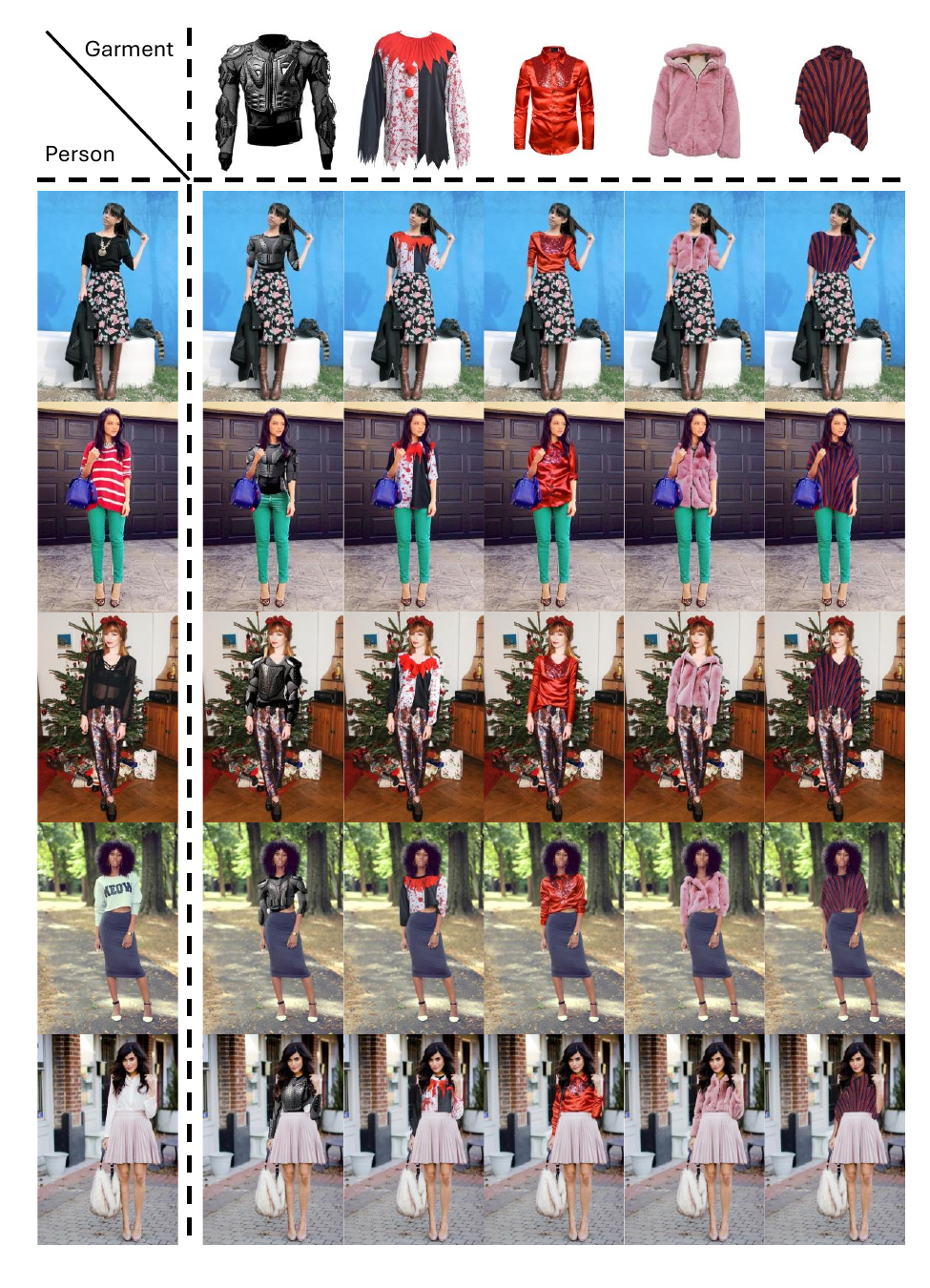}
    \caption{
    In-the-wild results generated by \ourmethod~ using a variety of reference garments beyond the native virtual try-on task. Best viewed when zoomed in.}
    \label{fig:results_grid_ood_refs}
\end{figure*}

\begin{figure*}[ht]
    \centering
    \includegraphics[width=0.95\textwidth]{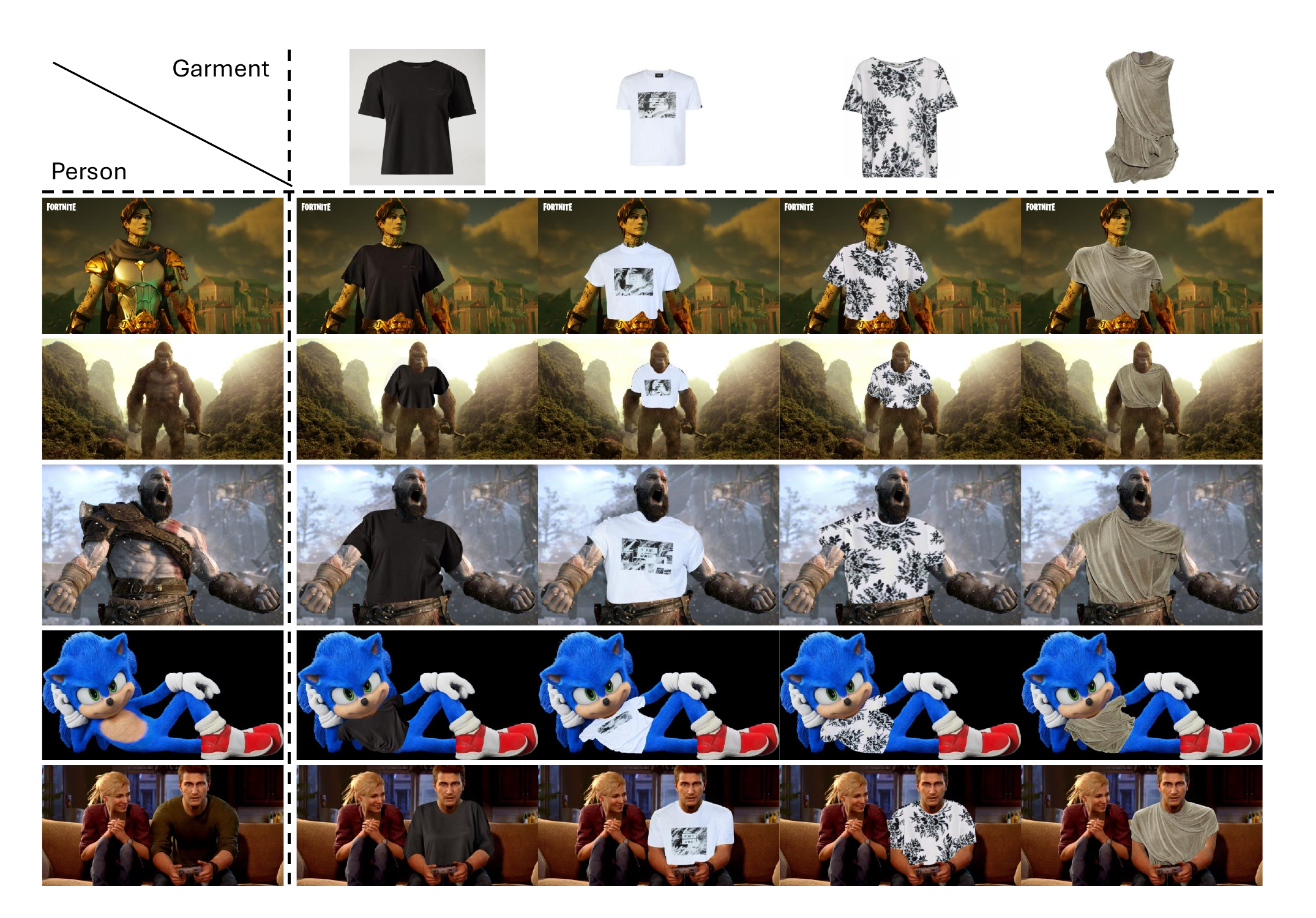}
    \caption{
    In-the-wild results generated by \ourmethod~ using unseen target person images which are out of the domain of the native virtual try-on task. Best viewed when zoomed in.}
    \label{fig:results_grid_ood_tars}
\end{figure*}

\begin{figure*}[ht]
    \centering
    \includegraphics[width=0.9\textwidth]{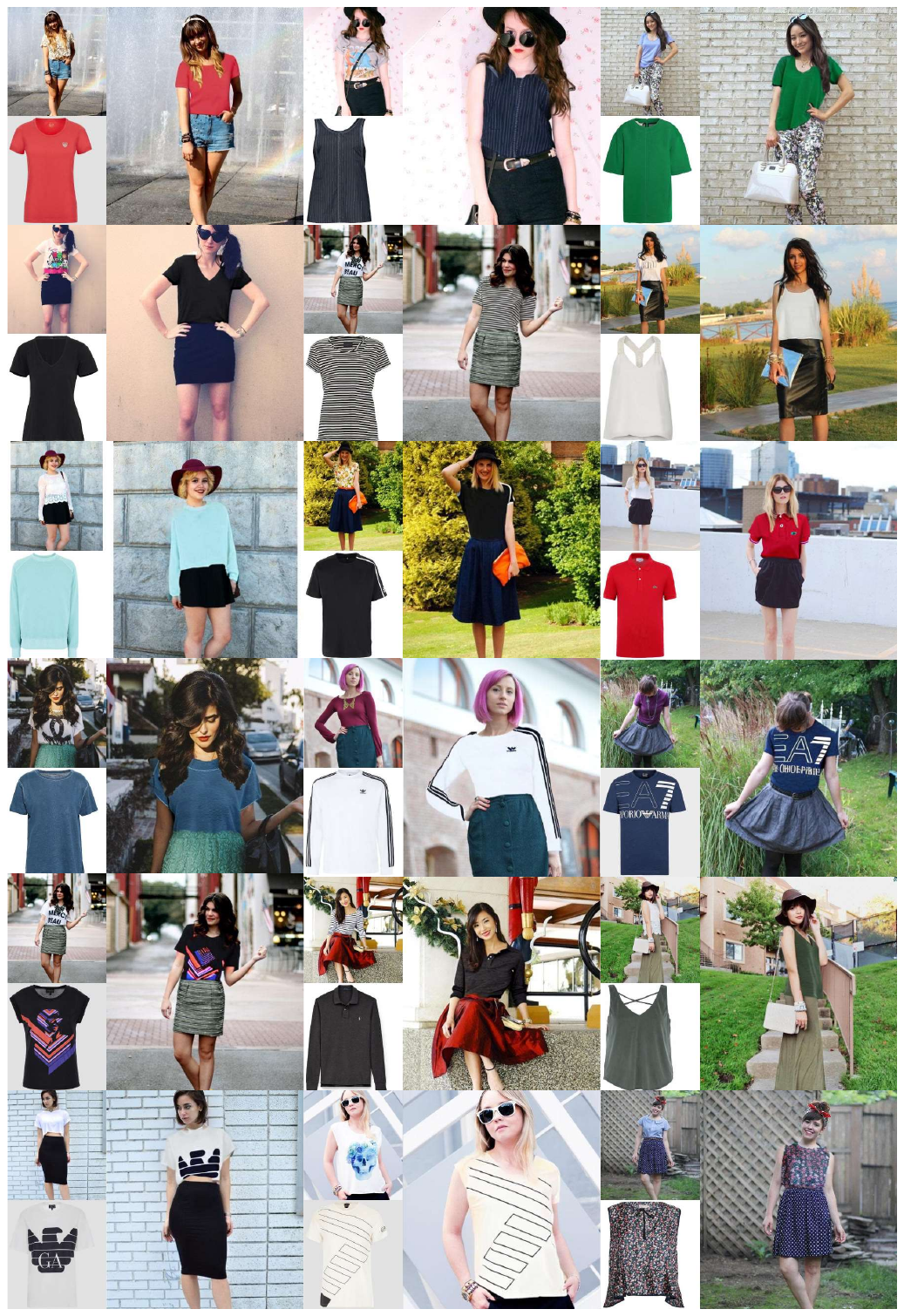}
    \caption{Results Gallery. Additional generated results by MaX4Zero. Best viewed when zoomed in.}
    \label{fig:sup1}
\end{figure*}

\begin{figure*}[ht]
    \centering
    \includegraphics[width=0.9\textwidth]{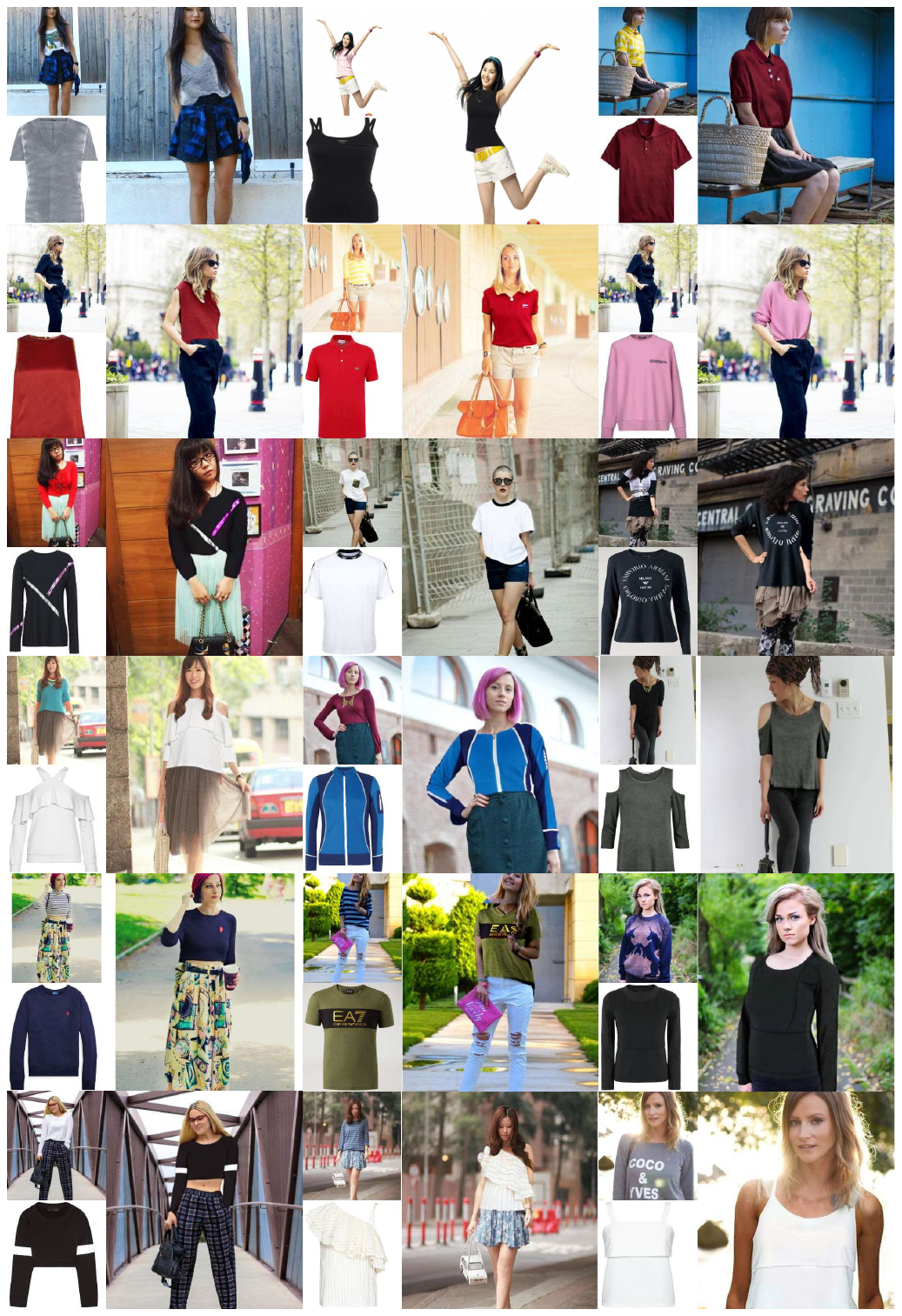}
    \caption{Results Gallery. Additional generated results by MaX4Zero. Best viewed when zoomed in.}
    \label{fig:sup2}
\end{figure*}

\begin{figure*}[ht]
    \centering
    \includegraphics[width=0.9\textwidth]{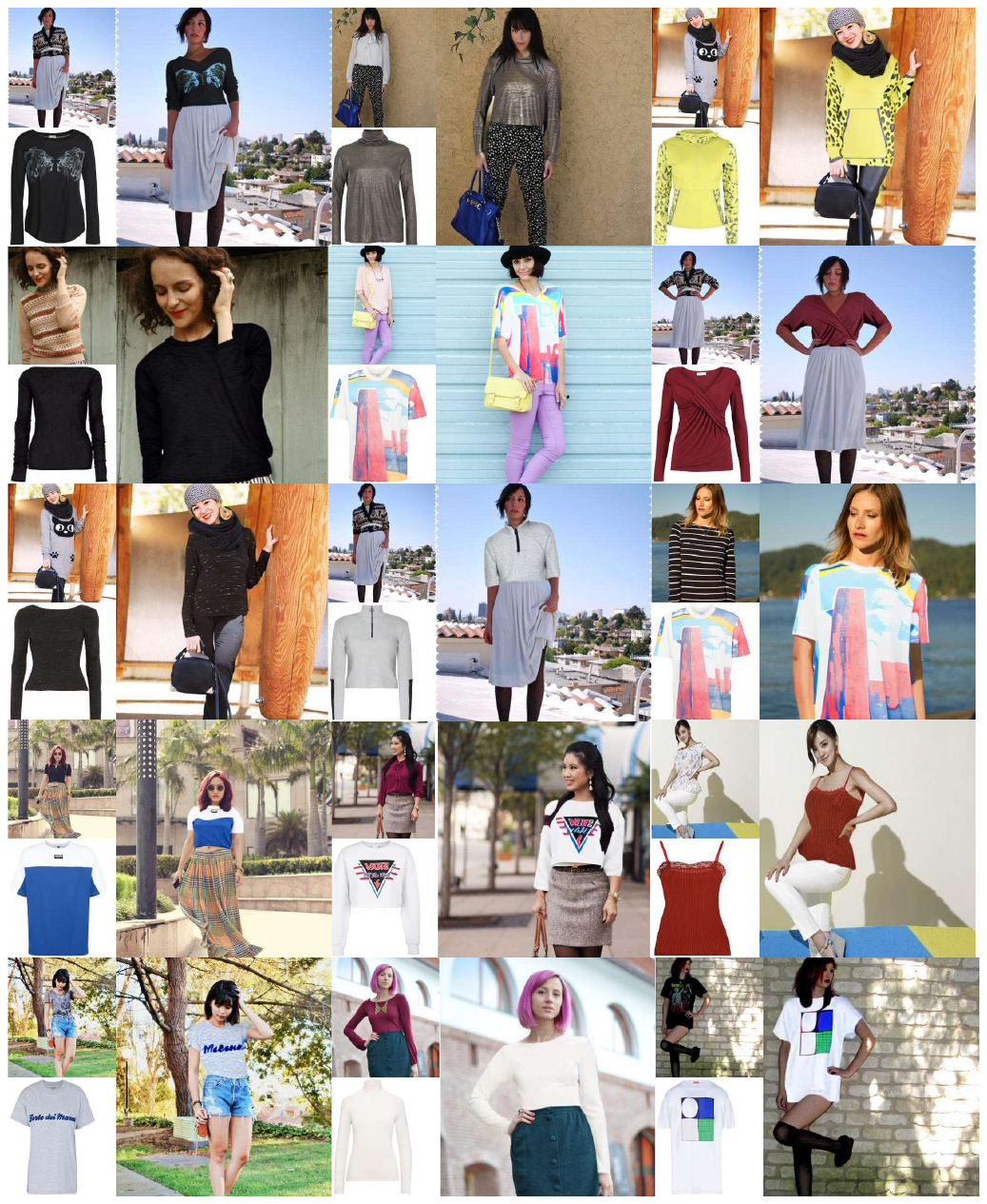}
    \caption{Results Gallery. Additional generated results by MaX4Zero. Best viewed when zoomed in.}
    \label{fig:sup3}
\end{figure*}

In the following figures, you will find additional qualitative results. Table-formatted results are shown in Fig. \ref{fig:results_grid}. Further results using out-of-domain reference garments or target figures can be found in Fig. \ref{fig:results_grid_ood_refs} and Fig. \ref{fig:results_grid_ood_tars}.  Supplementary results are available in Fig. \ref{fig:sup1}, Fig.\ref{fig:sup2} and Fig. \ref{fig:sup3}. These results, generated using mentioned hyperparameters, encompass examples featuring diverse human figures in natural settings and various garment types.

\end{document}